# Semantic Sentence Composition Reasoning for Multi-Hop Question Answering

**Qianglong Chen**[*]

College of Computer Science and Technology, Zhejiang University, Hangzhou, Zhejiang, 310000, China.
Email: chenqianglong@zju.edu.cn

**Abstract.** Due to the lack of insufficient data, existing multi-hop open domain question answering systems require to effectively find out relevant supporting facts according to each question. To alleviate the challenges of semantic factual sentences retrieval and multi-hop context expansion, we present a semantic sentence composition reasoning approach for multi-hop question answering task, which consists of two key modules: a multi-stage semantic matching module (MSSM) and a factual sentence composition module (FSC). With the combination of factual sentences and multi-stage semantic retrieval, our approach can provide more comprehensive contextual information for model training and reasoning. Experimental results demonstrate our model is able to incorporated with existing pre-trained language models and outperform existing SOTA method on the QASC task with an improvement of about 9%.

## 1. Introduction

Multi-hop reasoning for question answering is a huge challenge in question answering systems, which requires models to learn to reason over sentences or documents. Recent pre-trained language models [1-4] have gained impressive results on several question answering or machine comprehension datasets [5-7]. However, there are still challenges facing multi-hop question answering applications. We can classify these challenges into two categories: 1) retrieval challenges, i.e., finding the supporting factual sentence associated with each question. 2) reasoning challenges, i.e., learning compositional inference chains where the relations between entities appearing in retrieved factual sentences for the question are not obvious.

On one hand, existing works on NLP tasks with external knowledge graphs [8-13] have shown that semantic knowledge graphs, in particular, common sense and coreference information, can improve the performance of complex NLP models. However, most IR approaches often fail to reason over multi-hop facts which is critical for identifying right answer. There are few works focusing on combining semantic information retrieval models with natural language reasoning models. To alleviate above mentioned challenges, we present a semantic information retrieval approach, rather than just using BM25 scores implemented in Elasticsearch, to effectively retrieve facts.

On the other hand, there are semantic relations among retrieved facts for each question in common sense datasets. Reasoning models are therefore required to accomplish compositional reasoning over retrieved facts. For now, the Question Answering via Sentence Composition (QASC) dataset [14] has a collection of questions and multiple answer choices, which demands existing model retrieving factual evidence from a large evidence corpus and composing them for question answering inference

(As shown in figure 1). Models for QASC must be able to learn the compositions that may lead to the same semantic relation.

As the QASC dataset is the first large-scale dataset that has annotated the facts to be composed and the relations between the question and these factual sentences are not straightforward, we choose the QASC dataset to evaluate our model to handle retrieval and compositional reasoning challenges.

We summarize several contributions as follows:

1. To alleviate above retrieval challenges, a multi-stage semantic matching module (MSSM) is proposed instead of the standard IR systems and improve the performance of factual sentence retrieval on a large factual evidence corpus.

2. To address sentence composition challenge, we adopt a factual sentence composition module (FSC) to integrate the retrieved factual sentences for multi-hop reasoning.

3. Experimental results demonstrate that with the multi-stage semantic matching and factual sentence composition, our model can achieve new SOTA results on QASC task.

> Q1: A push forward is used for what by a spacecraft?
> Fact1: propulsion is used for flying by a spacecraft
> Fact2: Propulsion means to push forward or drive an object forward
> Answer 1: flying
> Q2: What is used to smooth decoupage?
> Fact1: sandpaper is used to smooth wooden objects
> Fact2: Traditionally, wooden objects are used in decoupage
> Answer 2: sandpaper

**Figure 1.** Examples in QASC dataset. Above examples show that the answer to the question must be inferred with at least two facts retrieved from a large-scale QASC corpus of factual sentences. Gold facts are not available in the test file.

## 2. The Related Works

*2.1. Open-Domain Multi-Hop QA*
To find potentially relative document, Chen et al. propose a TF-IDF based method [15] and infer the answer from obtained document. Meanwhile, some large-scale reading comprehension datasets require combining evidence from multiple sources [14,16,17]. Considering the importance of contextualized sentence-level representation, [18] propose an iterative multi-hop retrieval to retrieve the necessary evidence over a paragraph in the knowledge source to answer a question. Tu et al. [19] propose an interpretable SAE system which filters out the documents and reduce the distraction information by a document classifier.

*2.2. Semantic Information Retrieval*
In traditional approaches, Wu et al. [20] propose a vector space model with TF-IDF to compare difference of words in texts. However, existing traditional methods cannot capture the semantics information between individual words. Some works on re-ranking the retrieved paragraphs [21,22] aim to improve the retrieval of the TF-IDF based component.

We can divide existing neural network based semantic text matching models into representation-focused models [23,24] and interaction-focused models [25]. Considering that the retrieved facts can affect the performance directly, in this paper we treat semantic information retrieval as one of the most important parts in open-domain multi-hop QA.

*2.3. Semantic Compositional Sentence*

While a complex sentence can be divided into several components, we take the meaning of the sentence as the functional meaning of its components with combination rules [26]. While Zhu et al. [27] try to enhance LSTM models with prior sentimental knowledge, Qi et al. [28] integrate general linguistic knowledge in semantic composition models.

## 3. The Proposed Approach

As shown in figure 2, the framework of our approach consists of two novel modules and a standard open-domain QA architecture. To find out factual sentences relevant with the question, we first propose a multi-stage semantic matching module. Then we propose a factual sentences composition module to combine the factual sentences for latter multi-hop reasoning.

After these two modules, we employ a standard open-domain QA architecture for multi-choice prediction.

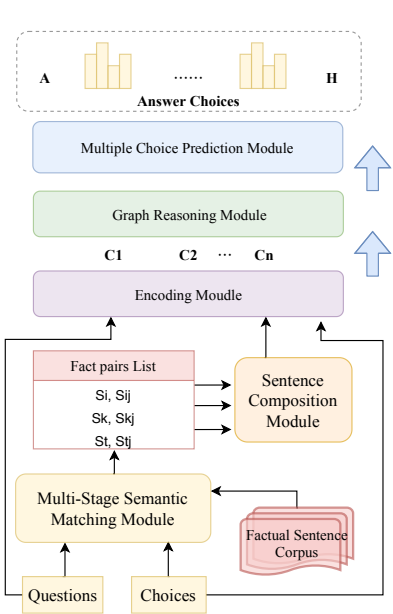
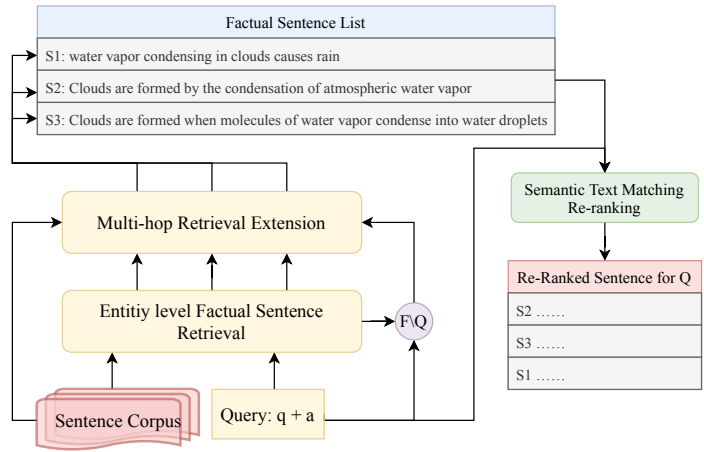

**Figure 2.** Workflow of our approach for multi-hop open domain question answering, two key modules are: (1) Multi-stage Semantic Matching, (2) Factual Sentence Composition.

**Figure 3.** Multi-Stage Semantic Matching Module. Q denotes the Query (question q and answer candidates a). $F\backslash Q$ denotes the difference set of entities between factual sentences and Query.

*3.1. Problem Formulation*

The open-domain QA task are formulated as $(Facts, Q, A)$ where Facts is a list of factual sentences retrieved from a large-scale corpus of factual sentences. The question $Q$ is a list of tokens, whose length is $n$. $A$ is a set of candidate answer choices, each of which is a series of tokens of length $m$. The objective is to find correct answer $a$ to given question $Q$ using relevant factual sentences, which can be formulated as $P(a|Facts, Q)$.

*3.2. Multi-Stage Semantic Matching Module*

As shown in figure 3, we first use a multi-stage information retrieval method to filter out the candidate knowledge facts from the large-scale corpus of factual sentences. Given the factual sentences corpus

$C_{all}$ and query $Q_i$ which contains question and answer choices tokens, we select the candidate pairs with Elasticsearch. In the entity-level factual sentence retrieval, we input $Q_i$ as the query and calculate rank scores between query entities and each factual sentence. As there are bridging entities in the different sentences, we calculate the difference set $D_i$ between the candidates and query entities and make another retrieval with entities in $D_i$ as expanded queries for multi-hop reasoning. After the initial and expanded retrievals, we use a semantic text matching model to re-rank those retrieved factual sentences.

*3.3. Factual Sentences Composition Module*

After getting top facts related to the question, we propose a factual sentences composition model for semantic reasoning.

Figure 4 shows that the retrieved factual sentences for the question contain exclusive entities and common entities. We refer to those common entities as bridging entities, which can be used by the composition module to consolidate the sentences of multiple stages into a complex sentence. The consolidated sentence usually contributes a lot to accomplish multi-hop inference properly.

In detail, we first identify the common words among the factual sentences. For example, we tokenize the facts into token level constituent and identify the common words *"is used to"*, *"be used in"* and *"wooden objects"* from facts *"sandpaper is used to smooth wooden objects"* *"traditionally, wooden objects are used in decoupage"*. Then, we filter the common words and concatenate the other words into composed sentence with heuristic rules.

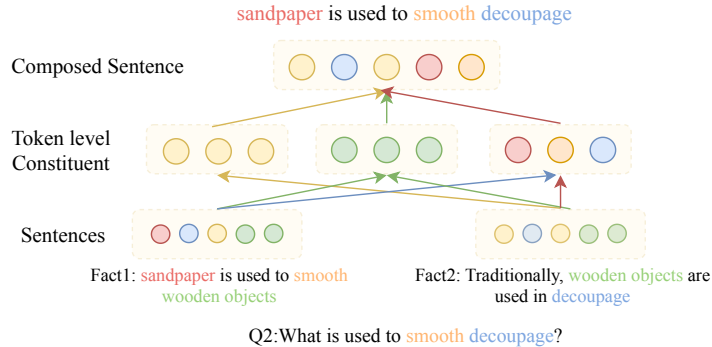

**Figure 4.** Factual Sentences Composition Module.

*3.4. Open-Domain QA Architecture*

The open-domain QA architecture consists of three parts: encoder module, reasoning module, and prediction module. In the encoder module, we concatenate composed factual sentences with the query $Q_i$ and then feed them into pre-trained RoBERTa or BERT, on top of which is a bi-attention layer. We denote the encoded representations as $C = \{c_1, c_2, \ldots, c_n\}$.

After encoding context, we adopt a Graph Attention Network (GAT) for graph reasoning, which considers all entity nodes as input and updates entity representation. After graph reasoning, the updated representations are exploited for multi-choice prediction. For loss function, we use cross-entropy during training.

**4. Main Experiments**

*4.1. QASC Dataset*

To empirically evaluate the proposed model, we leverage Question Answering via Sentence Composition (QASC) [14] for reasoning, which is a QA dataset with an emphasis on reasoning over compositional sentences. It consists of 9,980 8-way multiple-choice questions about primary school science (8,134 training examples, 926 development examples, 920 testing examples, respectively). The evidence corpus consists of 17M sentences.

*4.2. Experimental Setups*

We take pre-trained models, RoBERTa-base and BERT-wwm for parameters of the encoding layer initialization. For optimizer function, we adopt Adam. As for learning rate, we set it to 1e-5. Meanwhile, the number of training steps is 20000, warmup updates number is 150 and batch size is 8. For the metrics, we use the accuracy score for evaluation in our experiments.

*4.3. Baselines*

We use the two-step information retrieval method as one baseline. In this method, we first retrieve K facts $F_1$ based on the query $Q = q + a$, and for each $f_1 \in F_i$, retrieve another $L$ facts $F_2$, each of which contains at least one word appearing in $Q \backslash f_1$ or $f_1 \backslash Q$.

Other baselines we examined include notable pre-trained language models, as for example, BERT-large-cased (BERT LC).

*4.4. Results Analysis and Ablation Study*

As shown in Table 1, MSSM denotes multi-stage semantic matching module, MRE denotes multi-hop retrieval expansion module, and FSC denotes factual sentence composition module.

For open-domain question answering, the vanilla RoBERTa-base language model just performs 22.79%. Addition of entity-level IR to RoBERTa-base achieves the performance of 59.18%, which shows the importance of IR modules. Addition of multi-hop retrieval expansion module to RoBERTa-base performs 66.09%, indicating that MRE module can acquire factual sentences of higher quality than single-step and entity-level IR modules. Moreover, we compared the Two-step IR with multi-stage semantic matching. Experimental results show that multi-stage semantic matching module outperforms Two-step IR with an improvement of about 5% on the QASC development set.

Factual sentences composition module aims to consolidate multiple factual sentence pairs for better reasoning. Experimental results in table 1 show that with the combination of MSSM and FSC, RoBERTa-base (RoBERTa-base + MSSM +FSC) achieve the best performance of 81.52%. In other words, FSC module brings a further improvement of 4.6% on QASC development set.

Table 1. Accuracy Results of QASC dev set.

| Models | Dev Accuracy |
| --- | --- |
| RoBERTa-base w/o IR | 0.2279 |
| RoBERTa-base _+ Entity-level IR | 0.5918 |
| BERT LC + Single-step IR | 0.5982 |
| RoBERTa-base + MRE | 0.6609 |
| RoBERTa-base + Two-step IR | 0.7198 |
| RoBERTa-base + MSSM | 0.7692 |
| RoBERTa-base + MSSM +FSC | 0.8152 |

*4.5. Case Study*

For the question **"what is used to smooth decoupage"**, we retrieve the fact1 and fact2 (see figure 4) and compose these factual sentences into a short sentence where bridging entities are removed. Then the composed sentence and query are taken as input, going through graph reasoning and prediction modules, and generating the final output id $G$ whose token is "**sandpaper**".

**5. Conclusion**

To improve the retrieval results and multi-hop reasoning in question answering, we present a multi-stage semantic matching module and a factual sentences composition module, respectively. Experimental results fully demonstrate our modules outperform the standard IR system and two-step

IR method. We first combine the semantic sentence composition module with multi-stage semantic matching module, and it is suitable for any open-domain QA architecture. Further exploration will be done on the other question answering datasets with our proposed modules.

**References**
[1] Devlin J, Chang MW, Lee K, et al. 2019 BERT: Pre-training of Deep Bidirectional Transformers for Language Understanding *Proceedings of the 2019 Conference of the North American Chapter of the Association for Computational Linguistics: Human Language Technologies 1 (Long and Short Papers)*. (Minneapolis, Minnesota): 4171-4186.
[2] Liu Y, Ott M, Goyal N, et al. 2019 Roberta: A robustly optimized bert pretraining approach arXiv preprint arXiv:1907.11692
[3] Yang Z, Dai Z, Yang Y, et al. 2019 Xlnet: Generalized autoregressive pretraining for language understanding *33rd Conference on Neural Information Processing Systems*. (Vancouver, Canada): 5753-5763.
[4] Lan Z, Chen M, Goodman S, et al. 2020 Albert: A lite bert for self-supervised learning of language representations *Proceedings of International Conference on Learning Representations*. (Addis Ababa, Ethiopia). Online.
[5] Lai G, Xie Q, Liu H, et al. 2017 Race: Large-scale reading comprehension dataset from examinations *Proceedings of the 2017 Conference on Empirical Methods in Natural Language Processing*. (Copenhagen, Denmark): 785-794.
[6] Rajpurkar P, Jia R and Liang P 2018 Know what you don't know: Unanswerable questions for Squad *Proceedings of the 56th Annual Meeting of the Association for Computational Linguistics (Volume 2: Short Papers)*. (Melbourne, Australia): 784-789.
[7] Williams A, Nangia N and Bowman S R 2018 A broad-coverage challenge corpus for sentence understanding through inference *Proceedings of the 2018 Conference of the North American Chapter of the Association for Computational Linguistics: Human Language Technologies* 1 (Long Papers). (New Orleans, Louisiana): 1112-1122.
[8] Banerjee P, Pal K K, Mitra A, et al. 2019 Careful selection of knowledge to solve openbook question answering *Proceedings of the 57th Annual Meeting of the Association for Computational Linguistics*. (Florence, Italy): 6120-6129.
[9] Khot T, Sabharwal A and Clark P 2019 What's missing a knowledge gap guided approach for multi-hop question answering *Proceedings of the 2019 Conference on Empirical Methods in Natural Language Processing and the 9th International Joint Conference on Natural Language Processing*. (Hong Kong, China): 2814-2828.
[10] Jin D, Gao S, Kao J Y, et al. 2020 MMM: multi-stage multi-task learning for multi-choice reading comprehension *Proceedings of the AAAI Conference on Artificial Intelligence*. (New York, USA): 8010-8017.
[11] Moon S, Shah P, Kumar A, et al. 2019 OpenDialKG: Explainable Conversational Reasoning with Attention-based Walks over Knowledge Graphs *Proceedings of the 57th Annual Meeting of the Association for Computational Linguistics*. (Florence, Italy): 845-854.
[12] Yang P, Li L, Luo F, et al. 2019 Enhancing topic-to-essay generation with external commonsense knowledge *Proceedings of the 57th Annual Meeting of the Association for Computational Linguistics*. (Florence, Italy): 2002-2012.
[13] Yu J, Wang C, Luo G, et al. 2019 Course concept expansion in moocs with external knowledge and interactive game *Proceedings of the 57th Annual Meeting of the Association for Computational Linguistics*. (Florence, Italy): 4292-4302.
[14] Khot T, Clark P, Guerquin M, et al. 2020 QASC: A dataset for question answering via sentence composition *Proceedings of the AAAI Conference on Artificial Intelligence*. (New York, USA): 8082-8090.

[15] Chen D, Fisch A, Weston J, et al. 2017 Reading Wikipedia to answer open-domain question *Proceedings of the 55th Annual Meeting of the Association for Computational Linguistics (Volume 1: Long Papers)*. (Vancouver, Canada): 1870-1879.
[16] Talmor A, Berant J 2018 The web as a knowledge-base for answering complex questions *Proceedings of NAACL-HLT 2018*. (New Orleans, Louisiana): 641-651.
[17] Yang Z, Qi P, Zhang S, et al. 2018 Hotpotqa: A dataset for diverse, explainable multi-hop question answering *Proceedings of the 2018 Conference on Empirical Methods in Natural Language Processing*. (Brussels, Belgium): 2369-2380.
[18] Feldman Y and El-Yaniv R 2019 Multi-Hop Paragraph Retrieval for Open-Domain Question Answering *Proceedings of the 57th Annual Meeting of the Association for Computational Linguistics*. (Florence, Italy): 2296-2309.
[19] Tu M, Huang K, Wang G, et al. 2020 Select, answer and explain: Interpretable multi-hop reading comprehension over multiple documents *Proceedings of the AAAI Conference on Artificial Intelligence*. (New York, USA): 9073-9080.
[20] Wu H C, Luk R W P, Wong K F, et al. 2008 Interpreting tf-idf term weights as making relevance decisions *ACM Transactions on Information Systems (TOIS)*, **26**(3): 1-37.
[21] Htut P M, Bowman S R and Cho K 2018 Training a ranking function for open-domain question answering *Proceedings of the 2018 Conference of the North American Chapter of the Association for Computational Linguistics: Student Research Workshop*. (New Orleans, Louisiana, USA): 120-127.
[22] Yan M, Xia J, Wu C, et al. 2019 A deep cascade model for multi-document reading comprehension *Proceedings of the AAAI conference on artificial intelligence*. (Honolulu, Hawaii, USA): 7354-7361.
[23] Jiang J Y, Zhang M, Li C, et al. 2019 Semantic text matching for long-form documents *The World Wide Web Conference*. (San Francisco, CA, USA): 795-806.
[24] Wan S, Lan Y, Guo J, et al. 2016 A deep architecture for semantic matching with multiple positional sentence representations *Proceedings of the AAAI Conference on Artificial Intelligence*. (Phoenix, Arizona USA): 2835-2841.
[25] Dai Z, Xiong C, Callan J, et al. 2018 Convolutional neural networks for soft-matching n-grams in ad-hoc search *Proceedings of the eleventh ACM international conference on web search and data mining*. (Los Angeles, California, USA): 126-134.
[26] Pelletier F J 1994 The principle of semantic compositionality *Topoi*. **13(1)**: 11-24.
[27] Zhu X, Sobhani P and Guo H 2016 Dag-structured long short-term memory for semantic compositionality *Proceedings of the 2016 conference of the north American chapter of the association for computational linguistics: Human language technologies*. (San Diego, California): 917-926.
[28] Qi F, Huang J, Yang C, et al. 2019 Modelling semantic compositionality with sememe knowledge *Proceedings of the 57th Annual Meeting of the Association for Computational Linguistics*. (Florence, Italy):5706-5715.